\documentclass{article}
\usepackage{spconf,amsmath,graphicx,amssymb,makecell}
\usepackage[colorlinks,linkcolor=red]{hyperref}

\title{A Generative Adversarial Framework for Optimizing Image Matting and Harmonization Simultaneously}
\name{Xuqian Ren$^1$, Yifan Liu$^{2 \dagger }$\thanks{	$\dagger$ Corresponding author. E-mail: yifan.liu04@adelaide.edu.au}, Chunlei Song$^1$}
\address{$^1$School of Automation, Beijing Institute of Technology, Beijing, China\\ $^2$ School of Computer Science, The University of Adelaide, Adelaide, Australia}

\begin{document}
	\maketitle
	\begin{abstract}
		Image matting and image harmonization are two important tasks in image composition. Image matting, aiming to achieve foreground boundary details, and image harmonization, aiming to make the background compatible with the foreground, are both promising yet challenging tasks. Previous works consider optimizing these two tasks separately, which may lead to a sub-optimal solution. We propose to optimize matting and harmonization simultaneously to get better performance on both the two tasks and achieve more natural results. We propose a new Generative Adversarial (GAN) framework which optimizing the matting network and the harmonization network based on a self-attention discriminator. The discriminator is required to distinguish the natural images from different types of fake synthesis images. Extensive experiments on our constructed dataset demonstrate the effectiveness of our proposed method. Our dataset and dataset generating pipeline can be found in \url{https://git.io/HaMaGAN}.
	\end{abstract}
	\begin{keywords}
		image matting, image harmonization, generative adversarial, optimize simultaneously.
	\end{keywords}
	\section{Introduction}
	\label{sec:intro}
	
	Image composition, especially portrait composition, is a practical technique that can be applied to image editing, digital entertainment, advertisement, and so on. It requires a foreground matting based on the input portrait image, and then the foreground portrait is pasted another background image. However, since these two pictures are not taken simultaneously, the composite image will be unnatural. Therefore, image harmonization, which seeks to adjust the background to fit the foreground to make the whole picture look harmonious, is also important in the composition process. 
	
	Previous work has paid much attention to the image matting methods. ~\cite{xu2017deep,chen2018tom,chen2018semantic,wang2018deep,lutz2018alphagan,indexnet} focus on extracting foreground alpha mattes automatically. These methods use Adobe image dataset~\cite{rhemann2009perceptually} and have excellent results, but they often optimize the image matting problem on the alpha matte level and suppose that the new background image is unknown.
	
	Image harmonization methods such as ~\cite{zhu2015learning,tsai2017deep,cong2020dovenet} use the CNN model to transfer the background to be compatible with the foreground. Deep learning methods do better in results, but they often require a large number of paired images. However, there is usually no suitable composite and ground-truth pair of images for the image harmonization task. Cong.~\cite{cong2020dovenet} released a large-scale image harmonization dataset, which was generated by changing the foreground color segementated by coarse mask. They employ a network to adjust the foreground to harmonized with background used the coarse mask. It is not suitable for portrait editing, because it cannot separate the details, like the hair and fingers, which will bring difficulties to the harmonization. 
	
	Different from previous work, we consider to optimize the image matting and image harmonization simultaneously in one framework. Therefore the image matting may be more accurate and suitable for the new background, and the harmonization will also benefit from the more accurate input alpha matte.
	To achieve this goal, we conduct a new dataset for training these two tasks simultaneously. Inspired by~\cite{cong2020dovenet}, we can treat a real image as a harmonized one and segment the background region by alpha matte, rather than a rough mask. We adjust the background region to be inconsistent with the foreground, thus we can get a synthetic discordant composite image. The new dataset contains the portraits images from Matting Human Datasets ~\cite{Mattingdataset}. We name our new dataset as Human Matting and Harmonization dataset (HMH dataset).
	Besides, a new generative adversarial framework (GAN) is proposed for combining these two tasks together. 
	The framework has two generators; one is an image matting network, which is used to generate alpha mattes for the harmonization task, the other is an image harmonization network, which is used to process our composite image to make them harmonious. For quick implementation, we employ two off-the-shelf networks: the IndexNet~\cite{indexnet} for automatically generating the alpha matte and an attention U-Net~\cite{cong2020dovenet} for adjusting the color and details of the composite image. And we also use a self-attention discriminator~\cite{zhang2019self} to capture the non-local feature of the separated spatial region to help to optimize the IndexNet and the attention U-Net.
	
	To verify our new GAN framework's performance, we conducted several experiments on our constructed dataset. This work's main contributions are two-fold: 
	\begin{itemize}
		\item We propose a new GAN framework to optimize the image matting problem and image harmonization problem simultaneously. The proposed algorithm can improve the accuracy of the alpha mattes and optimize the details of the composite image.
		\item We release the first large-scale dataset, HMH dataset, for handling image harmonization and image matting for portraits together.
	\end{itemize}
	
	\section{METHODS}
	\label{sec:methods}
	\subsection{Overview}
	\label{sec:overview}
\begin{figure*}[t]
		\begin{center}
			\includegraphics[width=\linewidth]{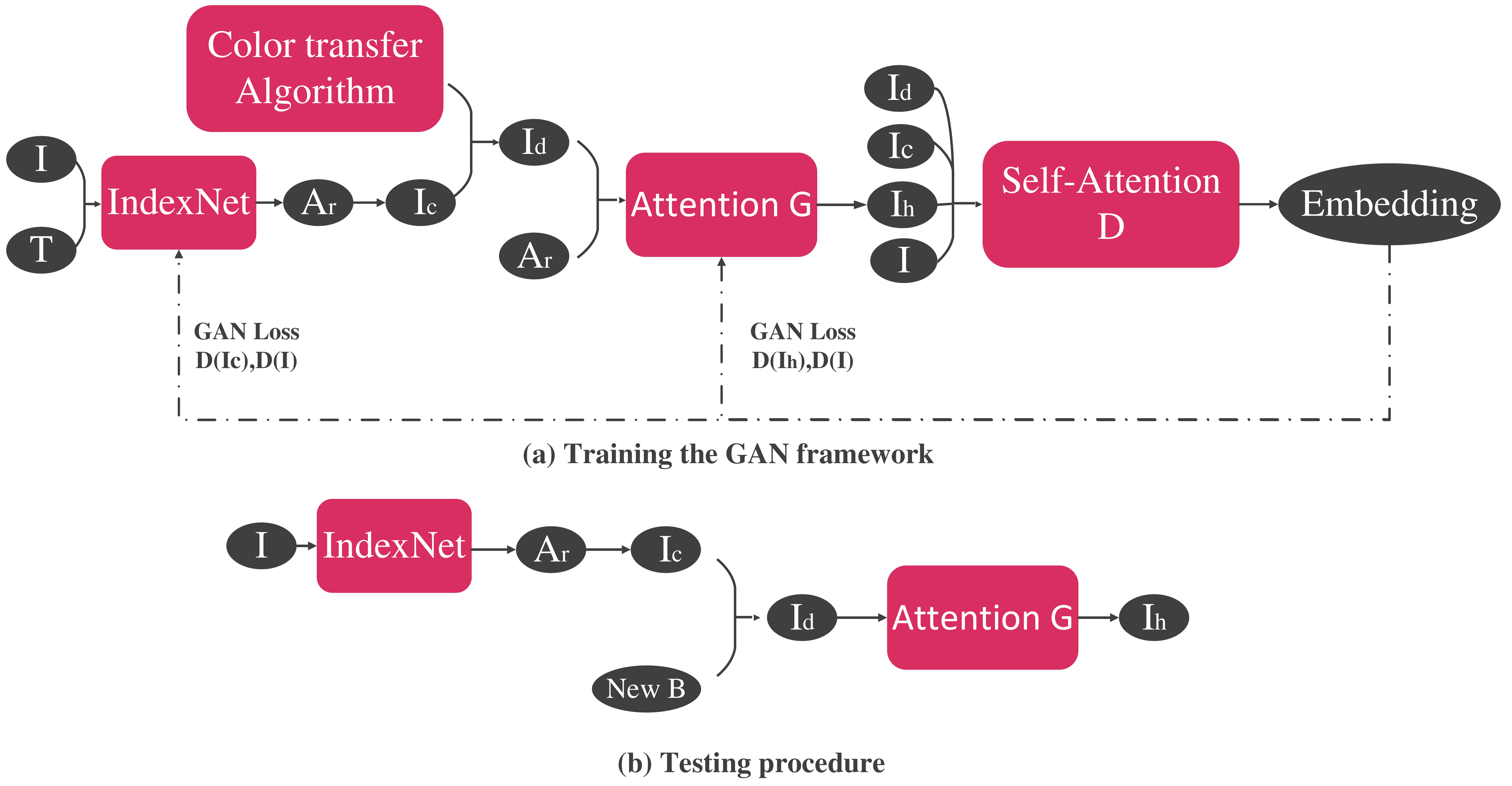}
		\end{center}
		\caption{The proposed pipeline of our framework. $I$: the real image, $T$: the trimap of the foreground, $A_r$:The result alpha matte generated by IndexNet~\cite{indexnet}, $I_c$: the composite image with fake foreground and original background, segmented by $A$, $I_d$: the disharmonious image which background has changed color or illumination, $I_h$: the harmonious image generated by attention G. new B: the new background selected randomly when testing.
		}
		\label{fig:structure}
		\vspace{-1em}
	\end{figure*}

	Our framework is based on generative adversarial networks (GANs), which can generate result alpha matte and harmonious image simultaneously. Figure~\ref{fig:structure} shows the structure of the proposed framework, which has three sub-networks. IndexNet~\cite{indexnet} is employed for the image matting task. An attention generator is used to perform the image harmonization task, and a self-attention discriminator is employed to optimize the two generators.
	
	The input of the IndexNet is a real image $I$ and its corresponding trimap $T$, and the output of the IndexNet is the image matting result alpha $A_{r}$. We train the IndexNet model to expect the result alpha matte $A_{r}$ to be as close to ground truth alpha matte as possible. 
	
	The harmonization network is added at the end of the image matting network. We composite a 
	disharmonious image $I_{d}$ with the predicted $A_{r}$, a synthetic background $B_f$ with color adjustment, and the original image as the foreground, 
	\begin{equation}
		{I_d} = {A_r}I + (1 - {A_r})B_f.
		\label{equation}
	\end{equation}
	The harmonization network takes $I_{d}$ and $A_r$ as inputs and produces the harmonious image $I_h$. A reconstructive loss between $I_h$ and $I$ is employed for the baseline.
	
	To make the generated image looks more natural, we further employ a discriminator to optimize these two tasks simultaneously. The discriminator is a self-attention~\cite{zhang2019self} model. We conduct several types of fake samples compare with the real sample $I$ to form the adversarial loss, as shown in Section~\ref{sec:D}.
	During the test phase, we choose to select a new background from the background dataset randomly and combine the new background with the predicted foreground split by $A_{r}$ from the original image as the input of the harmonization network and then get the final composite image.
	
	\subsection{Dataset Construction}\label{data_construction}
	\noindent{\bf Prepare Human Portrait Matting datasets:}  The original dataset we use is Matting Human Datasets~\cite{Mattingdataset} from Kaggle. The dataset has the real image $I$ and its corresponding alpha matte $A$. To make this dataset available to our framework, we use $A$ to segment out the foreground $F$ of $I$ and calculate the background $B$ through the inpaint function in OpenCv. Finally, we have four 
	sub-sets: $I$, $A$, $F$, $B$. 
	
	\noindent{\bf Background Adjustment:} In the background adjustment procedure, we randomly choose a transfer function to transfer the color, change the illumination degree, or enhance the color of the background image. In the procedure of the background adjustment, the color transfer method we use is Reinhard color transfer method~\cite{reinhard2001color} $R\left(  \cdot  \right)$. The color transfer function inputs are the background $B$ and a target image $I_t$ randomly selected from our training dataset, which is a different image that its background are different from the original background. The program extracts the target color from $I_t$ and transfers $ B $ and generates $B_f$. This process is shown below:
	\begin{equation}
		\begin{split}
			&{B_f} = R\left( {{B},{I_t}} \right)\\
		\end{split}
	\end{equation}
	Then $B_f$ will be recombined with the new foreground $F_f=A_{r}I$ as the disharmonious image $I_d$ as shown in equation~\ref{equation}. The process of the background adjustment are shown in Figure 3 in the supplementary materials.
	
	Finally, we get the training triplet for each image: an alpha matte, an RGB image, and a disharmonious image, which form our new Human Matting and Harmonization dataset (HMH dataset).
	
	\subsection{Framework}
	
	\noindent\textbf{{IndexNet Generator.}} We employ IndexNet as the image matting network. The IndexNet can dynamically predict indices for individual local regions, conditional on the input local feature map itself. More details can be referred in~\cite{indexnet}. We use alpha matte to generate trimap and mask, then we concatenate the real image and trimap and put them into the whole model. During training, we use the alpha prediction loss. Only loss from the unknown region of the trimap can be calculated. The output should be multiplied with mask $ M $, and the trimap should be multiplied with $\bar{M}=1-M$. Then they added to generate result alpha $A_{r}$. We want the $A_{r}$ to be close to real alpha image $A$ through $\ell_{G_{m_{mse}}}=\|A_{r}-A\|_{1}$. 
	
	\noindent\textbf{Attention Enhanced Generator.}	We employ a U-Net with attention blocks following~\cite{cong2020dovenet} as our harmonization network. Different from~\cite{cong2020dovenet}, we enforce the generated image $I_{h}=G_{h}(I_{d}, A_r)$ to be close to real image $I$ by $\ell_{G_{h_{mse}}}=\|I_{h}-I\|_{1}$.

	\noindent\textbf{Self-attention Discriminator}
	\label{sec:D}
	To generate a more natural composite image and optimize these two tasks together, we propose to use a self-attention discriminator~\cite{zhang2019self} to evaluate the difference between synthesis images and real images. And the generators are required to produce more natural results to mimic the distribution of real images.
	
	The discriminator is trained to distinguish the real images from several fake composite images:
	\begin{itemize}
		\item $I_c$. We use the predicted alpha matte to segment the foreground and composite it with the corresponding real background in the dataset.
		\item $I_d$. We use the predicted alpha matte to segment the foreground and composite it with the disharmonious background.
		\item $I_h$. 
		We use the prediction alpha matte to segment the foreground and composite it with the disharmonious background, and then the whole image is adjusted by the harmonization network to get a harmonious composite image.
	\end{itemize}
	
	Thus, the adversarial loss we used is,
	\begin{equation}
		\ell_D = \mathbb{E}[D({I_h}) +D({I_c})+D({I_d})- D(I)]
	\end{equation}
	
	When training, we need to minimize $\ell_D$, that is we want to produce large scores for real images and minimize the score of the generated image. The adversarial losses for the two generators are given by
	\begin{equation} 
		\begin{split}
			&\ell_{G_{m_D}} = \mathbb{E}[D(I)-D({I_c}) ]\\
			&{\ell_{G_{h_D}}} =  \mathbb{E}[D(I) - D({I_h})]
		\end{split}
	\end{equation}
	That is we want the generated images to fool the discriminator and obtain large scores. The total loss function for training the matting model $G_{m}$ and the harmonization model $G_h$ is:
	\begin{equation}
		\begin{split}
			&{\ell_{{G_m}}} = {\ell_{{G_{m_{mse}}}}} + \lambda_1 {\ell_{{G_{m_D}}}}\\
			&{\ell_{{G_h}}} = {\ell_{{G_{h_{mse}}}}} + \lambda_2 {\ell_{{G_{h_D}}}}
		\end{split}
	\end{equation}
	$\lambda_1$ and $\lambda_2$ are used to control the GAN loss weight, and we minimize ${\ell_{{G_m}}}$, ${\ell_{{G_h}}}$ to minimize the difference between the synthesis images and real images.
	
	\section{Experiments}
	\subsection{Implemention details}
	\noindent\textbf{Dataset and Evaluation Metric.} We perform our experiments on the conducted HMH dataset, which is build on Matting Human Datasets~\cite{Mattingdataset}. It is currently the largest portrait matting dataset, containing 34,427 images and corresponding alpha mattes. We split it into a training dataset with 30982 images and a testing dataset with 3444 images. During the test phase, the background dataset we use is the Scene UNderstanding (SUN) dataset~\cite{xiao2016sun}, which contains 130,519 images. All the image for training and testing are resized into ${\rm{256}} \times {\rm{256}}$. 
	
	For a fair comparison, we evaluate the image matting task using Mean Squared Error (MSE), Sum of Absolute Differences (SAD), perceptually motivated Gradient (Grad), and Connectivity (Conn) errors following~\cite{indexnet}. And we use Mean opinion score (MOS)~\cite{ledig2017photo} to evaluate the harmonized image results following~\cite{cong2020dovenet}.

	\noindent\textbf{Implemention Details.}	%
	For the matting network, we follow the training configurations used in~\cite{indexnet} and pretrianed the matting network on Adobe Image Dataset. For the harmonization network, we also use the configuration and structure in~\cite{cong2020dovenet} and pretrained the harmonization network on iHarmony4 dataset. 
	Then we train IndexNet~\cite{indexnet} on the conducted HMH dataset with the learning rate of $1e^{-7}$ for dconv, index and pred layers, and $1e^{-5}$ for other layers in fine-tuning stage. Finally, we jointly train the whole framework with the learning rate of 0.0002 for attention U-Net and self-attention discriminator, with the same learning rate in fine-tuning for IndexNet.
	
	\subsection{Results}
\begin{figure}[t]
		\begin{center}
			\includegraphics[width=\linewidth]{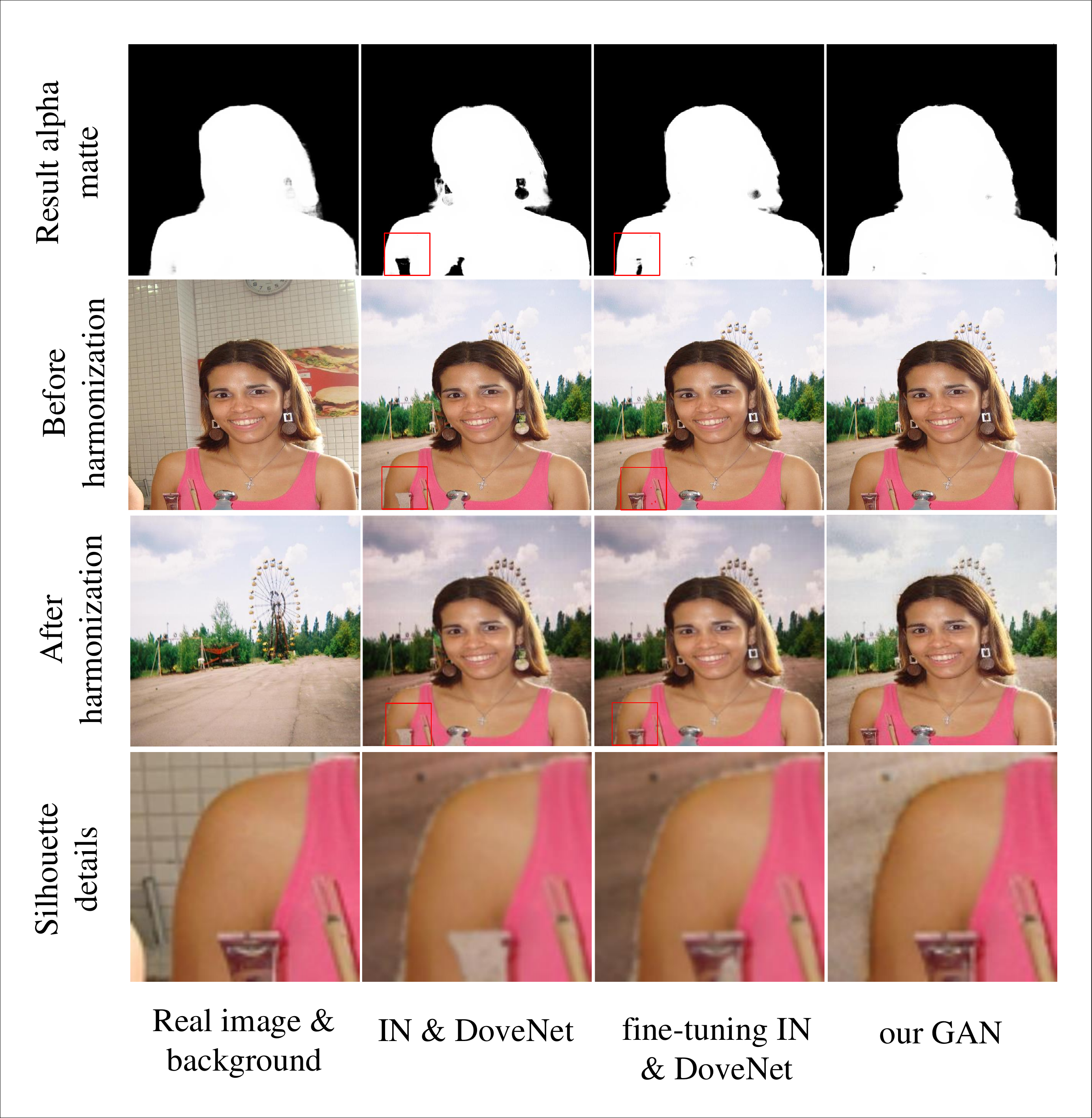}
		\end{center}
		\caption{This figure qualitatively illustrates the benefits of joint training. The first row shows the ground truth of one alpha matte and alpha results produced by pre-trained IndexNet, fine-tuning IndexNet, and our GAN framework. The second row shows the real image and the composite images with foreground segmented by the matting result from the previous row. The third row shows the background and the harmonization results processed by DoveNet or our GAN. The last row shows silhouette details of the model's shoulder of the images after different harmonization methods. The red box indicates that fine tuning and joint training can further improve the matting accuracy. The last row indicates that use a more accurate matte as an indicator can make a good transition between the foreground and the background.}
		\label{resultalpha}
		
	\end{figure}

	\noindent\textbf{Compared with Single Methods.}
	Previous models often work independently on image matting and image harmonization, leading to a sub-optimal. In Table~\ref{joint}, we show how the two tasks benefit from training in a GAN network. After IndexNet fine-tuned~\cite{indexnet} on our dataset, the MSE error has reduced $78.8\%$, which indicate there exits a domain gap between Adobe Image Dataset and our human matting dataset. We trained our GAN framework on our dataset with the parameters of $\lambda_1=0.02,\ \lambda_2=0.01$. Compare with the baseline IndexNet, jointly training can improve the results of matting with 11.69\% in MSE, 9.65\% in SAD, and 10.1\% in Conn. We also offer the three image matting qualitative results in Figure \ref{resultalpha}. We can see that the place circled by blue circles have holes on the matting result.
	
	To illustrate our method's effectiveness in the harmonization task, we also conducted a user study to quantify the harmonize images. Specifically, we asked 30 raters to assign an integral score from 1 (bad quality) to 5 (excellent quality) to the harmonized images presented randomly. Our scoring standard is that the background and foreground of pictures are coordinated and conform to the actual situation. The results have shown in Table~\ref{joint}. From the results, we can see that our method has a score of 3.582 in terms of MOS. We also show our qualitative harmonization results in Figure \ref{resultalpha}. It can be seen that there are obvious white edges in the two images in the middle of the last row, which indicates that a rough segmentation boundary will make the foreground and background of the harmonization image seem not integrated. Our method makes a good transition between the portrait and the background.
	
	\setlength{\tabcolsep}{1mm}{
		
		\begin{table}[h]
			\begin{center}
				\begin{tabular}{r|cccc|cc}
					\hline
					\makecell[r]{Task}& \multicolumn{4}{l|}{\makecell[c]{Matting}}&  \multicolumn{2}{l}{\makecell[c]{Harmonization}} \\ \hline
					Index   & MSE$\downarrow$    & SAD$\downarrow$    & Grad$\downarrow$    & Conn$\downarrow$    & {\makecell[c]{MOS$\uparrow$}} & {\makecell[c]{Std}}    \\\hline
					
					DM~\cite{xu2017deep}$^\ddag$ &0.65  &70.93& 29.81& 16.26 & \multicolumn{2}{l}{\makecell[c]{-}}              \\
					IN~\cite{indexnet} &0.11&16.43&15.05& 16.78  & \multicolumn{2}{l}{\makecell[c]{-}}              \\ 
					IN~\cite{indexnet}$^\ddag$ & 0.02 & 4.39 & \textbf{8.65}  & 4.28  & \multicolumn{2}{l}{\makecell[c]{-}}              \\ \hline
					Xue~\cite{xue2012understanding} & \multicolumn{4}{l|}{\makecell[c]{-}}            & \makecell[c]{2.32}  &  \textbf{0.40}  \\
					DIH~\cite{tsai2017deep} & \multicolumn{4}{l|}{\makecell[c]{-}}            & \makecell[c]{2.72}   & 0.47   \\
					U-Net~\cite{zhu2015learning} & \multicolumn{4}{l|}{\makecell[c]{-}}      & \makecell[c]{3.30}  & 0.54  \\
					DN~\cite{cong2020dovenet} & \multicolumn{4}{l|}{\makecell[c]{-}}            & \makecell[c]{3.15}  & 0.58  \\ \hline
					Our          & \textbf{0.02} & \textbf{3.96} & 8.68 & \textbf{3.84} & \makecell[c]{\textbf{3.58}} & 0.55 \\ \hline
				\end{tabular}
			\end{center}
			\caption{Image matting and harmonization results with or without GAN. The two tasks can benefit from jointly training. $\ddag$ means we fine-tune the network on our dataset. Std is the standard deviation of MOS.}  
			
			\label{joint}
		\end{table}
	}

	\noindent\textbf{Compared with Existing Image Harmonization Methods.}
	We also compare our model with exsiting image harmonization methods~\cite{xue2012understanding,tsai2017deep,cong2020dovenet} qualitatively, and show the results in Figure 4 in supplemental material.
	From Figure 4, we can see that our method can produce more natural composite images compared with other methods.

	\noindent\textbf{Image Matting Result with Different $\lambda_1$.} This section tests our image matting effect with different $\lambda_1$ parameters and with the same parameter $\lambda_2=0.01$. Then we compared our results alpha mattes with the ground truth alpha mattes pixel by pixel and shown the MSE results in Table~\ref{alpha}. We can see that increasing the $\lambda_1$ value can improve the matting precision in a specific range. When $\lambda_1=0.02$, it has the minimum amount of MSE at present. 
	\begin{table}[h]
		\setlength{\tabcolsep}{3mm}
		\begin{center}
			\begin{tabular}{l|cccc}
				\hline
				Method & MSE$\downarrow$&SAD$\downarrow$&Grad$\downarrow$&Conn$\downarrow$\\
				\hline
				$\lambda_1=0.01$ & 0.022&4.104&9.170&3.996 \\
				
				$\lambda_1=0.02$&\textbf{0.020} &\textbf{3.962}&\textbf{8.681}&\textbf{3.843}\\
				
				$\lambda_1=0.033$& 0.022&4.086&9.206&3.987\\
				$\lambda_1=0.1$ &0.022 &4.289&9.077&4.166\\
				\hline
			\end{tabular}
		\end{center}
		\caption{MSE results of image matting. In a certain range, with the increase of $\lambda_1$, it can make the precision better.}
		\label{alpha}
	\end{table}
	
	\noindent\textbf{Image Harmonization Result with Different $\lambda_2$.}
	We also test image harmonious effect with different $\lambda_2$, and with the same parameter $\lambda_1=0.01$. the qualitative image can be seen in Figure 5 in supplemental material. We can see as the parameter $\lambda_2$ increases, the background tends to change to a color closer to the foreground color and tends to become darker.

	\noindent\textbf{Input Foreground Alpha or Background Alpha Matte.}
	We also do some experiments to adjust the color of foreground to meet with the background using the same methods. Here Figure 6 in supplemental material shows our algorithm's effect sample. In some pictures, the faces of the portrait are very red and do not match the background. Our algorithm can produce more natural results when changing the foreground.

	\section{Conclusions}
	In this work, we have proposed a new GAN framework to optimize the image matting model and image harmonization model simultaneously, and an original dataset composite method to generate a dataset that can be used both for two tasks. From experiments, we can confirm that joint training can generate better alpha matte as well as more a realistic harmonization effect. Our method shows the feasibility of simultaneously optimizing image matting and image harmonization tasks. In the future, we can use some training techniques and data enhancement methods further to improve the training accuracy and the effect of the framework.

	\newpage
	{
		\bibliographystyle{IEEEbib}
		\bibliography{arxiv}
	}
	
	\section{Supplementary Material}
	 In this section, we show our supplementary images.
	\begin{figure*}[t]
		\begin{center}
			\includegraphics[width=\linewidth]{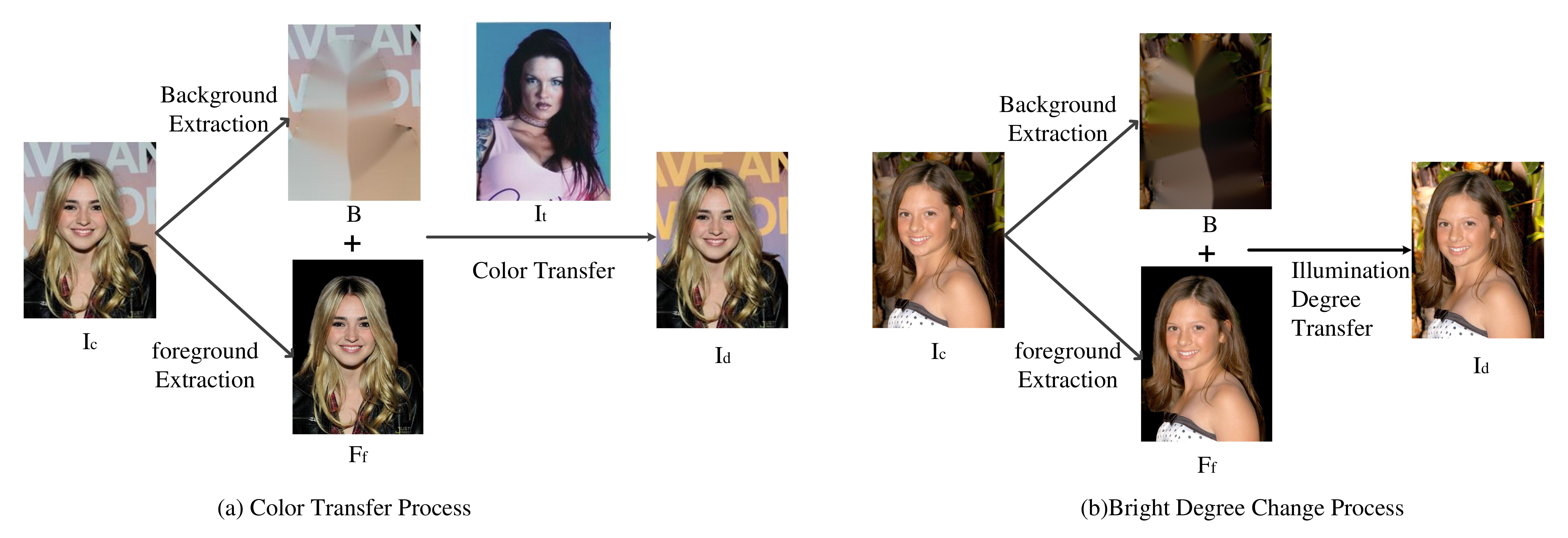}
		\end{center}
		\caption{This figure illustrates our color transfer and illumination adjustment procedures. In the color transfer procedure, we first use result alpha matte $A_r$ to segment the fake foreground $F_f$, then we transfer the color of our origin background $B$ to the target color extracted from $I_t$, finally, we composite $F_f$ and fake background with color adjustment using $A_r$. In the illumination transfer process, we also first extract the Ff, then change the bright degree of the original background, then we combine the fake background with a different bright degree with the fake foreground as the disharmonious image $I_d$.}
		\label{3}
		
	\end{figure*}
	
	\begin{figure*}[t]
		\begin{center}
			\includegraphics[width=\linewidth]{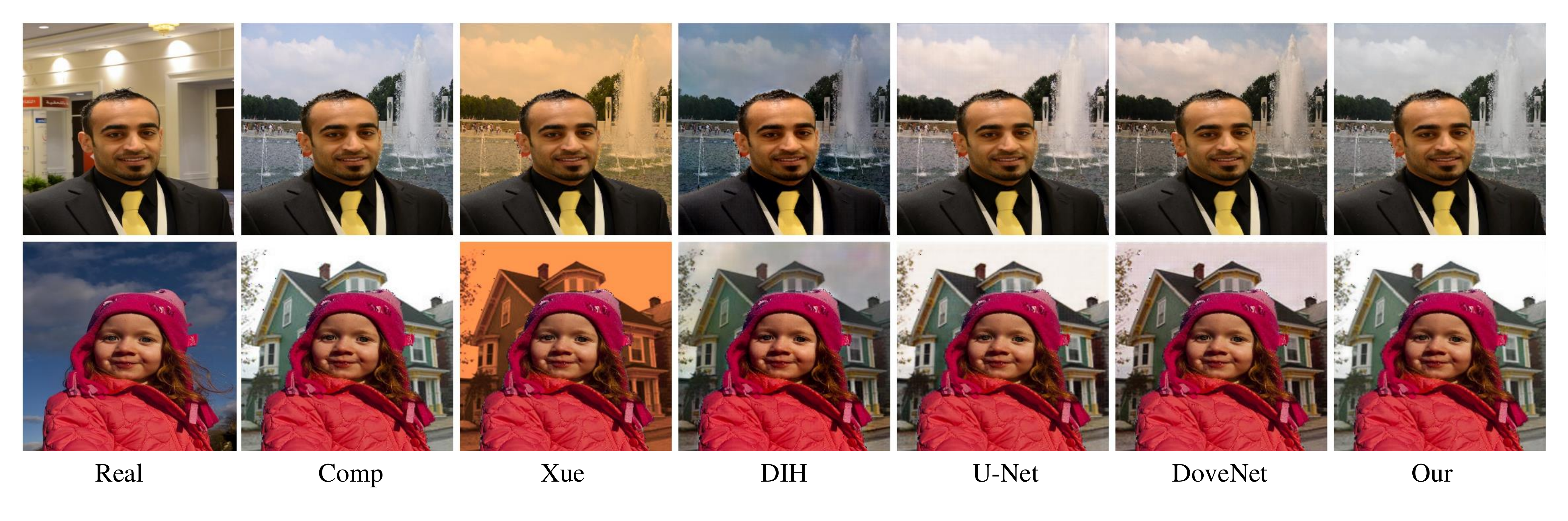}
		\end{center}
		\caption{This figure shows example results of different methods on our dataset. From top to bottom, we show three examples from our dataset. We provide the composite image, Xue, DIH, U-net+attention, DoveNet and our method results on our dataset, from left to right.}
		\label{4}
	\end{figure*}
	
	\begin{figure}[t]
		\begin{center}
			\includegraphics[width=\linewidth]{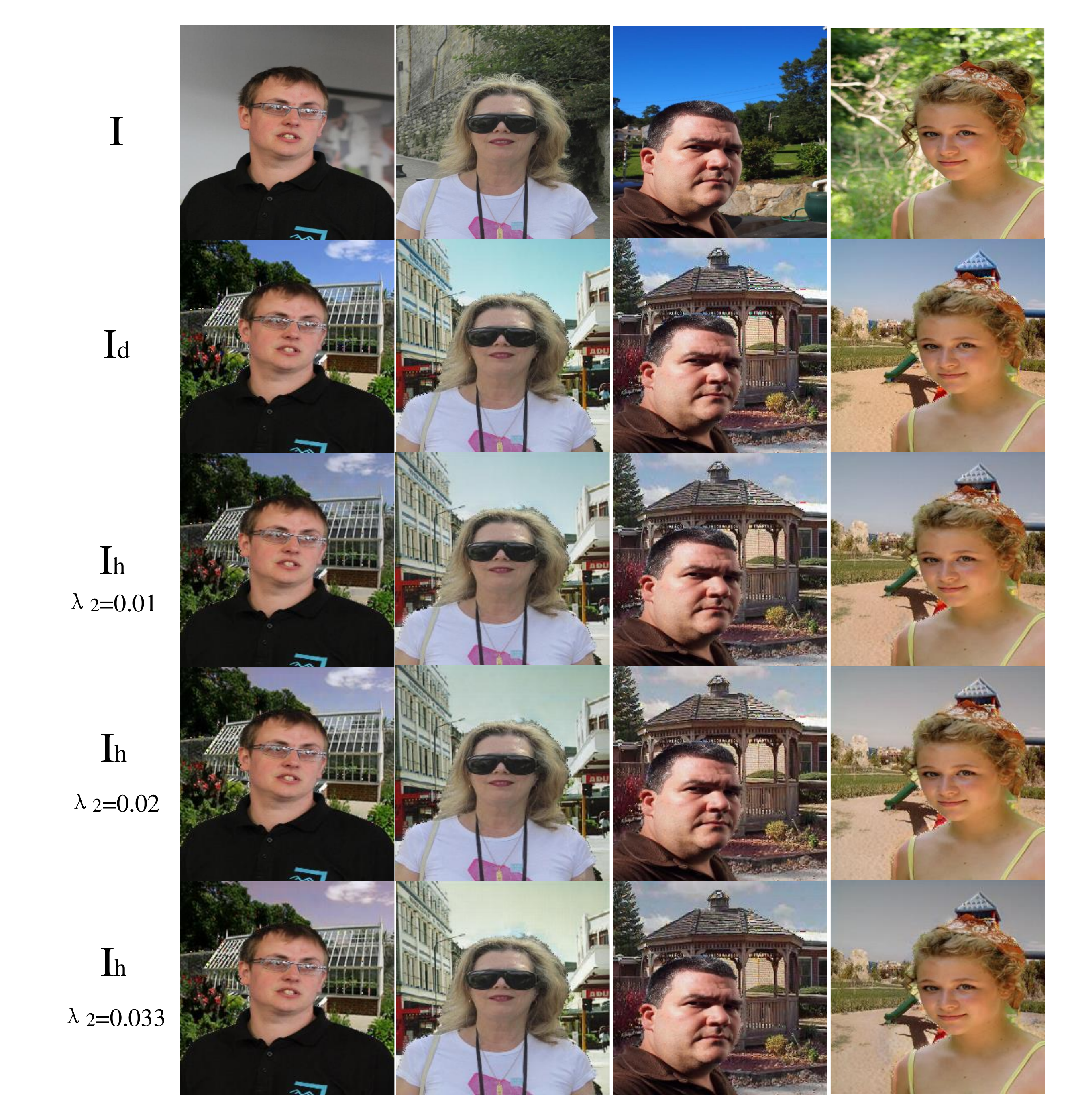}
		\end{center}
		\caption{This figure shows qualitative harmonization effect with different $\lambda_2$. We can see that with the increase of $\lambda$, the background tends to change to a color closer to the foreground color.}
		\label{5}
	\end{figure}
	
	\begin{figure}[t]
		\begin{center}
			\includegraphics[width=\linewidth]{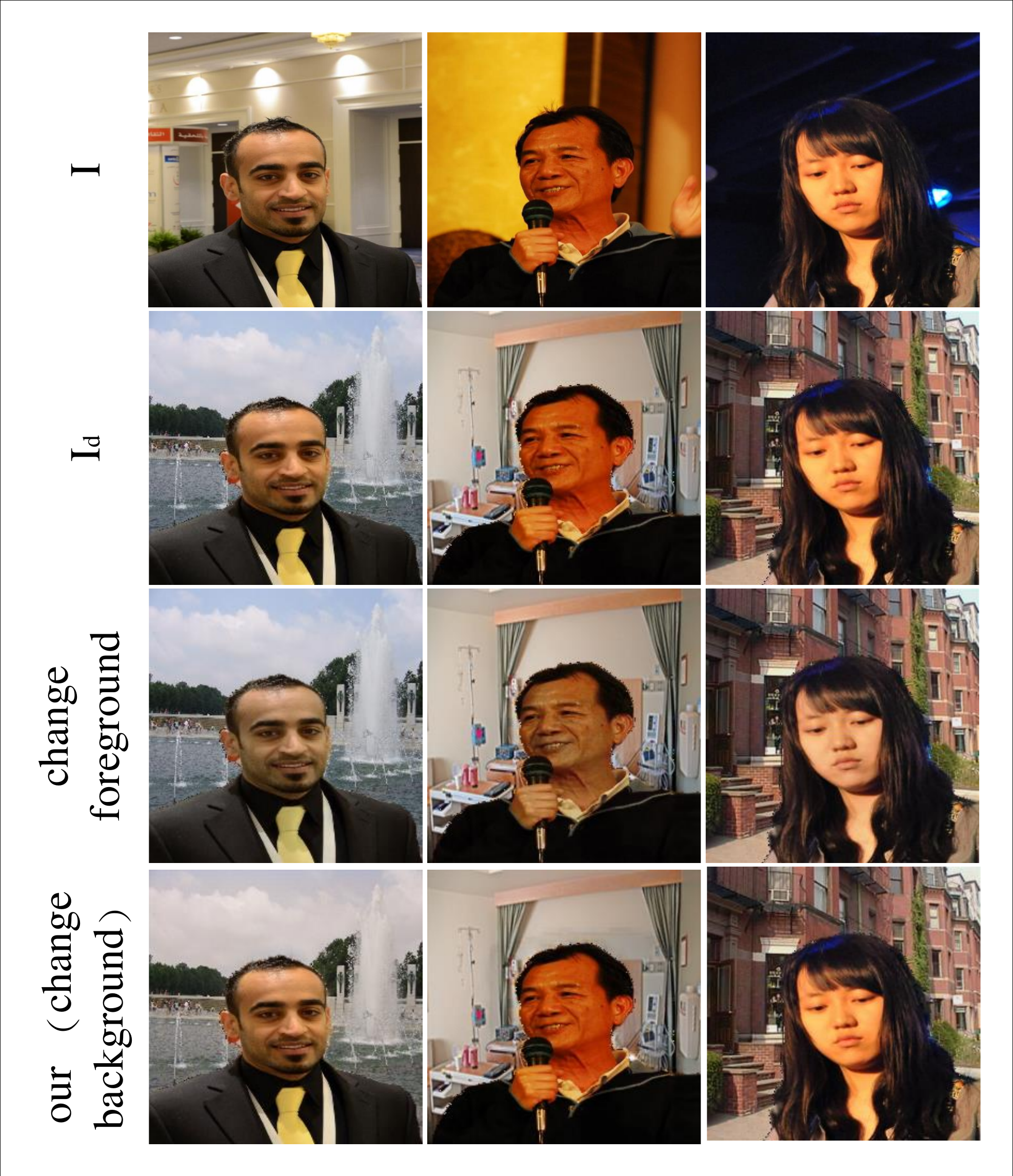}
		\end{center}
		\caption{This figure shows the effect of our algorithm on changing an image's foreground color to adapt to the background. We can see a red color in the portrait face has reduced and matches the background.}
		\label{6}
	\end{figure}

\end{document}